\documentclass[runningheads]{llncs}
\usepackage{amsmath,amssymb} %
\usepackage[utf8]{inputenc}
\usepackage{subfigure}
\usepackage{graphicx}
\usepackage{comment}

\usepackage{color}
\usepackage{tabularx}
\usepackage{booktabs} %

\newcommand\sL{\ensuremath{\mathcal{L}}}

\newcommand\sX{\ensuremath{\mathcal{X}}}
\newcommand\sY{\ensuremath{\mathcal{Y}}}

\newcommand\inv[1]{\ensuremath{\frac{1}{#1}}}

\newcommand\eqdef{\ensuremath{\stackrel{\rm def}{=}}} %

\usepackage{tikz}
\usetikzlibrary{fit,positioning,bayesnet}

\usepackage{xcolor}
\newcommand{\HB}[1]{{\color{red}{HB: #1}}} %
\renewcommand{\HB}[1]{\ignorespaces}
\newcommand{\etal}{et al.}
\usepackage[ruled,vlined]{algorithm2e}
\usepackage{algorithmic}

\usepackage{bm}
\usepackage{wrapfig}
\usepackage{adjustbox}%

\begin{document}
\pagestyle{headings}
\mainmatter
\def\ECCVSubNumber{4034}  %

\title{AutoSimulate: (Quickly) Learning Synthetic Data Generation} %

\titlerunning{AutoSimulate: (Quickly) Learning Synthetic Data Generation}
\author{Harkirat Singh Behl\inst{1} \and
Atilim Güneş Baydin\inst{1} \and
Ran Gal\inst{2} \and
Philip H.S. Torr\inst{1} \and
Vibhav Vineet\inst{2}
}
\index{Behl, Harkirat Singh}
\index{Baydin, Atilim Gunes}
\authorrunning{Behl et al.}
\institute{Univeristy of Oxford, Oxford, UK\\
\email{\{harkirat,gunes,phst\}@robots.ox.ac.uk}
\and
Microsoft Research, Redmond, USA \\
\email{\{rgal,vibhav.vineet\}@microsoft.com}\\
}

\maketitle

\begin{abstract}
Simulation is increasingly being used for generating large labelled datasets in many machine learning problems. Recent methods have focused on adjusting simulator parameters with the goal of maximising accuracy on a validation task, usually relying on REINFORCE-like gradient estimators. However these approaches are very expensive as they treat the entire data generation, model training, and validation pipeline as a black-box and require multiple costly objective evaluations at each iteration. We propose an efficient alternative for optimal synthetic data generation, based on a novel differentiable approximation of the objective. This allows us to optimize the simulator, which may be non-differentiable, requiring only one objective evaluation at each iteration with a little overhead. We demonstrate on a state-of-the-art photorealistic renderer that the proposed method finds the optimal data distribution faster (up to $50\times$), with significantly reduced training data generation (up to $30\times$) and better accuracy ($+8.7\%$) on real-world test datasets than previous methods.

\keywords{synthetic data, training data distribution, simulator, optimization, rendering}
\end{abstract}

\section{Introduction}

Massive amounts of data needs to be collected and labelled for training neural networks for tasks such as object detection \cite{ren2017faster,he2017mask}, segmentation \cite{Shelhamer_2017} and machine translation \cite{LuongPM15}.
A tantalizing alternative to real data for training neural networks has been the use of synthetic data, which provides accurate labels for many computer vision and machine learning tasks such as (dense) optical flow estimation \cite{Dosovitskiy_2015_ICCV,RichterVRK_eccv16}, pose estimation \cite{Varol_2017,doersch2019sim2real,XiangSNF_corr18,TekinSF_corr17,KehlMTIN_iccv17}, among others \cite{SuQLG15,gaidon2016virtual,hinterstoisser2017pre,RadL_iccv17,dwibedi2017cut,RosSMVL_cvpr16}.
Current paradigm for synthetic data generation involves human experts manually handcrafting the distributions over simulator parameters \cite{le_2016_synthetic_data,Sakaridis_2018}, or randomizing the parameters to synthesize large amounts of data using game engines or photorealistic renderers \cite{Dosovitskiy_2015_ICCV,RichterVRK_eccv16,RosSMVL_cvpr16}.
However, photorealistic data generation with these approaches is expensive, needs significant human effort and expertise, and can be sub-optimal. 
This raises the question, has the full potential of synthetic data really been utilized?

\begin{figure}[t]
    \centering
    \includegraphics[width=\textwidth]{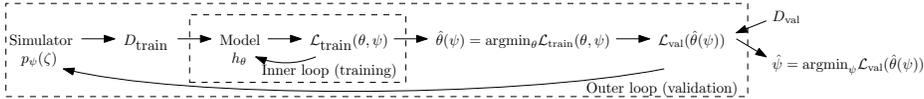}
    \caption{\textbf{Overview of the bilevel optimization setup.} A simulator $p_{\bm{\psi}}(\zeta)$ is used to generate a synthetic dataset $D_{\textrm{train}}$; \textbf{inner loop}: a model $h_{\bm{\theta}}$ is then trained on this dataset with training loss $\mathcal{L}_{\text{train}}(\bm{\theta}, \bm{\psi})$ to obtain optimal model paramaters $\hat{\bm{\theta}}(\bm{\psi})$; a real-data validation set $D_{\textrm{val}}$ is used to evaluate the performance of this trained model with validation loss $\mathcal{L}_{\textrm{val}}(\hat{\bm{\theta}}(\bm{\psi}))$, providing a measure of goodness of simulator parameter $\bm{\psi}$; \textbf{outer loop}: $\bm{\psi}$ is updated until we find optimal simulator parameters $\hat{\bm{\psi}}$. \HB{change this figure}}
    \label{fig:problem_figure}
\end{figure}

Recent approaches \cite{louppe2017adversarial,Ganin2018SynthesizingPF,ruiz2018learning,kar2019metasim} have formulated the setting of simulator parameters as a learning problem. A few of these methods \cite{louppe2017adversarial,Ganin2018SynthesizingPF} learn simulator parameters to minimize the distance between distributions of simulated data and real data. Ruiz \etal \cite{ruiz2018learning} proposed to learn the optimal simulator parameters to directly maximise the accuracy of a model on a defined task. 
However these approaches \cite{kar2019metasim,ruiz2018learning} are very expensive, as they treat the entire data generation and model training pipeline (Figure~\ref{fig:problem_figure}, outer loop) as a black-box, and use policy gradients \cite{Williams_1992}, which require multiple expensive objective evaluations at each iteration.
As a result, learning synthetic data generation with photorealistic renderers has remained a challenge.

In this work, we propose a fast optimization algorithm for learning synthetic data generation, which can quickly optimize state-of-the-art photorealistic renderers.
We look at the problem of finding optimal simulator parameters as a bi-level optimization problem (Figure~\ref{fig:problem_figure}) of training (inner) and validation (outer) iterations, and derive approximations for their corresponding objectives. Our key contribution lies in proposing a novel differentiable approximation of the objective, which allows us to optimize the simulator requiring only one objective evaluation at each iteration, with improved speed and accuracy. We also propose effective numerical techniques to optimize the approximation, which can be used to derive terms depending on desired speed-accuracy tradeoff.
The proposed method can be used with non-differentiable simulators and handle very deep neural networks. We demonstrate our method on two renderers, the Clevr data generator \cite{clevr_cite} and the state-of-the-art photorealistic renderer Arnold \cite{georgiev2018arnold}. \HB{write about real world data}

\section{Related Work}
\subsubsection{Expert Involvement and Random Generation}
One of the initial successful work on training deep neural networks on synthetic data for a computer vision problem was done on optical flow estimation, where 
Dosovitskiy et al. \cite{Dosovitskiy_2015_ICCV} created a large dataset by randomly generating images of chairs using an OpenGL pipeline by pasting objects onto randomly selected real-world images. This strategy has been applied in other problems like object detection, instance segmentation and pose estimation \cite{SuQLG15,hinterstoisser2017pre,RadL_iccv17,TekinSF_corr17,dwibedi2017cut}. Though this approach is simple to implement, the foreground objects are always pasted onto out-of-context background images, thereby requiring careful selection of the background images to achieve good accuracy as shown by Dvornik et al. \cite{dvornik2018}. Other issues with this technique include these images not being realistic, objects not having accurate shading, and shadows being inconsistent with the background.

Another line of work explores generation of photorealistic images with objects rendered within complete 3D scenes \cite{RichterHK_iccv17,RichterVRK_eccv16,hodan2019photorealistic,RosSMVL_cvpr16,gaidon2016virtual,TremblayTB_corr18,HandaPBSC_corr15a,ZhangSYSLJF_cvpr17}. Though this is a well accepted approach for synthetic data generation, it suffers from several issues. First, given the data generation process is independent of the neural network training, these approaches synthesize a large set of redundant training images, as shown in experiments section. This might add a massive redundant burden on the rendering infrastructure. 
Second, this requires human expert involvement, e.g., to set the right scene properties, material and texture of objects, quality of rendering, among several other simulator parameters \cite{hodan2019photorealistic}. This hinders widespread adaptation of synthetic data to different tasks. Finally, some sub-optimal synthesized data can corrupt the neural network training.

\subsubsection{Learning Simulator Parameters}
In order to resolve these issues, recent research has focused on learning the simulator parameters. The non-differentiability of the simulators has posed a challenge for optimization. Louppe et al. \cite{louppe2017adversarial} proposed an adversarial variational optimization technique for learning parameters of non-differentiable simulators, by minimizing the Jensen--Shannon divergence between the distribution of the synthetic data and distribution of the real data. Ganin et al. \cite{Ganin2018SynthesizingPF} incorporated a non-differentiable simulator within an adversarial training pipeline for generating realistic synthetic images.

Ruiz et al. \cite{ruiz2018learning} focused on optimization of simulator parameters with the objective of generating data that directly maximizes accuracy on downstream tasks such as object detection. They treated the entire pipeline of data generation and neural network training as a black-box, and used classical REINFORCE-based \cite{Williams_1992} gradient estimation.
However, this approach suffers from scalability issues. A single objective evaluation involves generating a synthetic dataset, training a neural network for multiple epochs, and calculating the validation loss. And this method requires mutiple such expensive objective evaluations for taking a single step.
Thus it has a very slow convergence and is difficult to scale to photorealistic simulators which have hundreds of parameters. In contrast, our method AutoSimulate, requires only a single objective evaluation at each iteration and works well with state-of-the-art photorealistic renderers.
Making an assumption that a probabilistic grammar is available, Kar et al. \cite{kar2019metasim} proposed to learn to transform the scene graphs within this probabilistic grammar, with the objective of simulataneously optimizing performance on downstream task and matching the distribution of synthetic images to real images. They also use REINFORCE-based \cite{Williams_1992} gradient estimation for the first objective like \cite{ruiz2018learning}, whose limitations were discussed above. 

In bi-level optimization, differentiating through neural network training is a challenge. \cite{mackay2019self} proposed to learn an approximation of inner loop using another network. Concurrent work \cite{yang2019learning} makes an assumption that the neural network is trained only for one or few iterations (not epochs), so they can store the computation graph in memory and back-propagate the derivatives, in a similar spirit as MAML \cite{finn2017model,behl-2019-alphamaml}. In constrast, we proposed a novel differentiable approximation of the inner loop using a Newton step which can handle many epochs without memory constraints. We also proposed efficient approximations which can be used for desired speed-accuracy tradeoff.

\section{Problem Formulation}
In supervised learning, a training set $D_{\textrm{train}}=\{z_1,...,z_m\}$ of input--output pairs $z_i = (\bm{x}_i, \bm{y}_i) \in \sX \times \sY$ is used to learn the parameters $\bm{\theta} \in \mathbb{R}^n$ of a model $h_{\bm{\theta}}$ that maps the input domain $\sX$ to the output codomain $\sY$. This is accomplished by minimizing the empirical risk $\frac{1}{m} \sum_{i=1}^{m} l(z_i, \bm{\theta})$, where $l(z, \bm{\theta}) \in \mathbb{R}$ denotes the loss of model $h_{\bm{\theta}}$ on a data point $z$.

Our goal is to generate synthetic training data using a simulator such that the model trained on this data minimizes the empirical risk on some real-data validation set $D_{\text{val}}$.
The simulator defines a data generating distribution $p_{\bm{\psi}}(\zeta)$ given simulator parameters $\bm{\psi} \in \mathbb{R}^m$, from which we can sample training data instances $\zeta \sim p_{\bm{\psi}}(\zeta)$, where we use $\zeta$ to denote simulated data as opposed to real data $z$.
The objective of finding optimal simulator parameters $\hat{\bm{\psi}}$ can then be formulated as the optimization problem
\begin{subequations}
\begin{align}
\min_{\bm{\psi}} \quad	&\sL_{\text{val}}\big(\hat{\bm{\theta}}(\bm{\psi})\big)\label{eq:main_obj_psi} &\\
s.t. \quad              &\hat{\bm{\theta}}(\bm{\psi}) \in \arg\min_{\bm{\theta}} \sL_{\text{train}}(\bm{\theta}, \bm{\psi})\;,\label{eq:main_obj_theta}              
\end{align}
\end{subequations}
where $\sL_{\text{val}}\big(\hat{\bm{\theta}}(\bm{\psi})\big) = \sum_{z_i \in D_{\text{val}}} l\big(z_i, \hat{\bm\theta}(\bm{\psi})\big)$ is the validation loss, $\sL_{\text{train}}(\bm{\theta}, \bm{\psi}) = \mathop{\mathbb{E}}_{\zeta \sim p_{\bm{\psi}}} \big[l(\zeta, \bm{\theta})\big]$ is the training loss, $\hat{\bm\theta}(\bm{\psi})$ denote the optimum of model parameters after training on data generated from the simulator parameterised by $\bm{\psi}$, and $\hat{\bm{\psi}}$ denote the optimum simulator parameters that minimize $\sL_{\textrm{val}}$. In this paper we will refer to Equations~\ref{eq:main_obj_psi} and \ref{eq:main_obj_theta} as the outer and inner optimization problems respectively. This formulation is illustrated in Figure~\ref{fig:problem_figure}.

Equations~\ref{eq:main_obj_psi} and \ref{eq:main_obj_theta} represent a bi-level optimization problem \cite{bilevel,franceschi2018bilevel,bennett2008bilevel}, which is a special kind of optimization where one problem is nested within another.
To compute the gradient of the objective $\sL_{\text{val}}\big(\hat{\bm\theta}(\bm{\psi})\big)$ with respect to $\bm{\psi}$, one needs to propagate derivatives through the training of a model and data generation from a simulator, which is often impossible due to the simulator being non-differentiable \cite{louppe2017adversarial}. Even in the case of a differentiable simulator, backpropagating through entire training sessions is impracticable because it requires keeping a large number of intermediate variables in memory \cite{maclaurin2015gradient}. %
One technique to address this challenge is to treat the entire system as a black-box and use off-the-shelf hyper-parameter optimization algorithms such as REINFORCE \cite{Williams_1992}, evolutionary algorithms \cite{mitchell1998introduction} or Bayesian optimization \cite{snoek2012practical}, which require multiple costly evaluations of the objective in each iteration. An important distinction from neural network hyper-parameter optimization is that evaluating the objective $\sL_{\text{val}}\big(\hat{\bm\theta}(\bm{\psi})\big)$ at a given $\bm{\psi}$ is much more expensive in our setting because it involves the expensive step of running the simulation for synthetic dataset generation along with neural network training.

In this paper we propose an efficient technique based on locally approximating the objective function $\sL_{\text{val}}\big(\hat{\bm\theta}(\bm{\psi})\big)$ at a point $\bm{\psi}$, together with an effective numerical procedure to optimize this local model, enabling the efficient tuning of simulator parameters in state-of-the-art computer vision workflows. %

\section{AutoSimulate}

We will derive differentiable approximations of the outer and inner optimization problems (Figure~\ref{fig:problem_figure}) using Taylor expansions of the objectives $\sL_{\text{val}}$ and $\sL_{\text{train}}$.

\textbf{Outer Problem} Our goal is to find $\hat{\bm{\psi}}$, the optimal simulator parameters which minimise $\sL_{\text{val}}(\hat{\bm{\theta}}(\bm{\psi}))$ in the outer (validation) problem, so we construct a Taylor expansion of $\sL_{\text{val}}(\hat{\bm{\theta}}(\bm{\psi}))$ around $\bm{\psi}_t$ at iteration $t$ as
\begin{align} \label{eq:taylor_val}
\sL_{\text{val}}\big(\hat{\bm{\theta}}(\bm{\psi}_t + \Delta\bm{\psi})\big) &= \sL_{\text{val}}\big(\hat{\bm{\theta}}(\bm{\psi}_t)\big) + \Delta\bm{\psi} \frac{d\hat{\bm{\theta}}(\bm{\psi}_t)}{d\bm{\psi}} \frac{d \sL_{\text{val}}\big(\hat{\bm{\theta}}(\bm{\psi}_t)\big)}{d\hat{\bm{\theta}}(\bm{\psi}_t)}  + ... \nonumber\\
&= \sL_{\text{val}}\big(\hat{\bm{\theta}}(\bm{\psi}_t)\big) + \Delta\hat{\bm{\theta}}_{\bm{\psi}} \frac{d \sL_{\text{val}}\big(\hat{\bm{\theta}}(\bm{\psi}_t)\big)}{d\hat{\bm{\theta}}(\bm{\psi}_t)} + ...\;,
\end{align}
where $\Delta\hat{\bm{\theta}}_{\bm{\psi}} = \Delta\bm{\psi} \frac{d\hat{\bm{\theta}}(\bm{\psi}_t)}{d\bm{\psi}} \approx \hat{\bm{\theta}}(\bm{\psi}_t + d\bm{\psi}) - \hat{\bm{\theta}}(\bm{\psi}_t)$.

\textbf{Inner Problem} To obtain parameter update $\Delta\hat{\bm{\theta}}_{\bm{\psi}}$ for the inner (training) problem, which requires retraining on the dataset generated with the new simulator parameter $\bm{\psi}_t + \Delta\bm{\psi}$, we write the loss function $\sL_{\text{train}}(\bm{\theta}, \bm{\psi}_t + \Delta\bm{\psi})$ as its Taylor series approximation around the current $\hat{\bm{\theta}}(\bm{\psi})$ as
\begin{align} \label{eq:taylor_train}
\sL_{\text{train}}\big(\hat{\bm{\theta}}(\bm{\psi}_t) + \Delta\bm{\theta}, \bm{\psi}_t + \Delta\bm{\psi}\big) 
&= \sL_{\text{train}}\big(\hat{\bm{\theta}}(\bm{\psi}_t), \bm{\psi}_t + \Delta\bm{\psi}\big) \nonumber\\ 
&+ \Delta\bm{\theta}^\top \frac{\partial}{\partial\bm{\theta}} \sL_{\text{train}}\big(\hat{\bm{\theta}}(\bm{\psi}_t), \bm{\psi}_t+ \Delta\bm{\psi}\big) \nonumber\\
&+ \frac{1}{2}\Delta\bm{\theta}^\top \bm{H}\big(\hat{\bm{\theta}}(\bm{\psi}_t), \bm{\psi}_t+ \Delta\bm{\psi}\big) \Delta\bm{\theta} + ...\; ,
\end{align}
where the Hessian $\bm{H}\big(\hat{\bm{\theta}}(\bm{\psi}_t), \bm{\psi}_t+ \Delta\bm{\psi}\big) \eqdef \frac{\partial^2}{\partial\bm{\theta}^2} \sL_{\text{train}}\big(\hat{\bm{\theta}}(\bm{\psi}_t), \bm{\psi}_t+ \Delta\bm{\psi}\big) \in \mathbb{R}^{n\times n}$.

We are interested in our local model in the limit $\Delta\bm{\psi} \to 0$,
implying that our initial point $\hat{\bm{\theta}}(\bm{\psi}_t)$ will be in close vicinity to the optimal $\hat{\bm{\theta}}(\bm{\psi}_t + \Delta\bm{\psi})$. Thus we utilize the local convergence of the Newton method \cite{nocedal2006numerical} and approximate $\sL_{\text{train}}(\bm{\theta}, \bm{\psi}_t + \Delta\bm{\psi})$ by the quadratic portion. Assuming the $\bm{H}\big(\hat{\bm{\theta}}(\bm{\psi}_t), \bm{\psi}_t+ \Delta\bm{\psi}\big)$ is positive definite, and minimizing the quadratic portion with respect to $\Delta\bm{\theta}$, we get
\begin{align}
\Delta\hat{\bm{\theta}}_{\bm{\psi}} &\approx \arg\min_{\Delta\bm{\theta}} \Big(\Delta\bm{\theta}\!^\top \frac{\partial \sL_{\text{train}}\big(\hat{\bm{\theta}}(\bm{\psi}_t), \bm{\psi}_t+ \Delta\bm{\psi}\big)}{\partial\bm{\theta}} \!+ \!\frac{1}{2}\Delta\bm{\theta}\!^\top \bm{H}\big(\hat{\bm{\theta}}(\bm{\psi}_t), \bm{\psi}_t+ \Delta\bm{\psi}\big) \Delta\bm{\theta}\Big) \nonumber\\
&= - \bm{H}\big(\hat{\bm{\theta}}(\bm{\psi}_t), \bm{\psi}_t + \Delta\bm{\psi}\big)^{-1} \frac{\partial \sL_{\text{train}}\big(\hat{\bm{\theta}}(\bm{\psi}_t), \bm{\psi}_t+ \Delta\bm{\psi}\big)}{\partial\bm{\theta}}\;.
\end{align}

\begin{figure}
    \centering
    \includegraphics[height=0.5\textwidth]{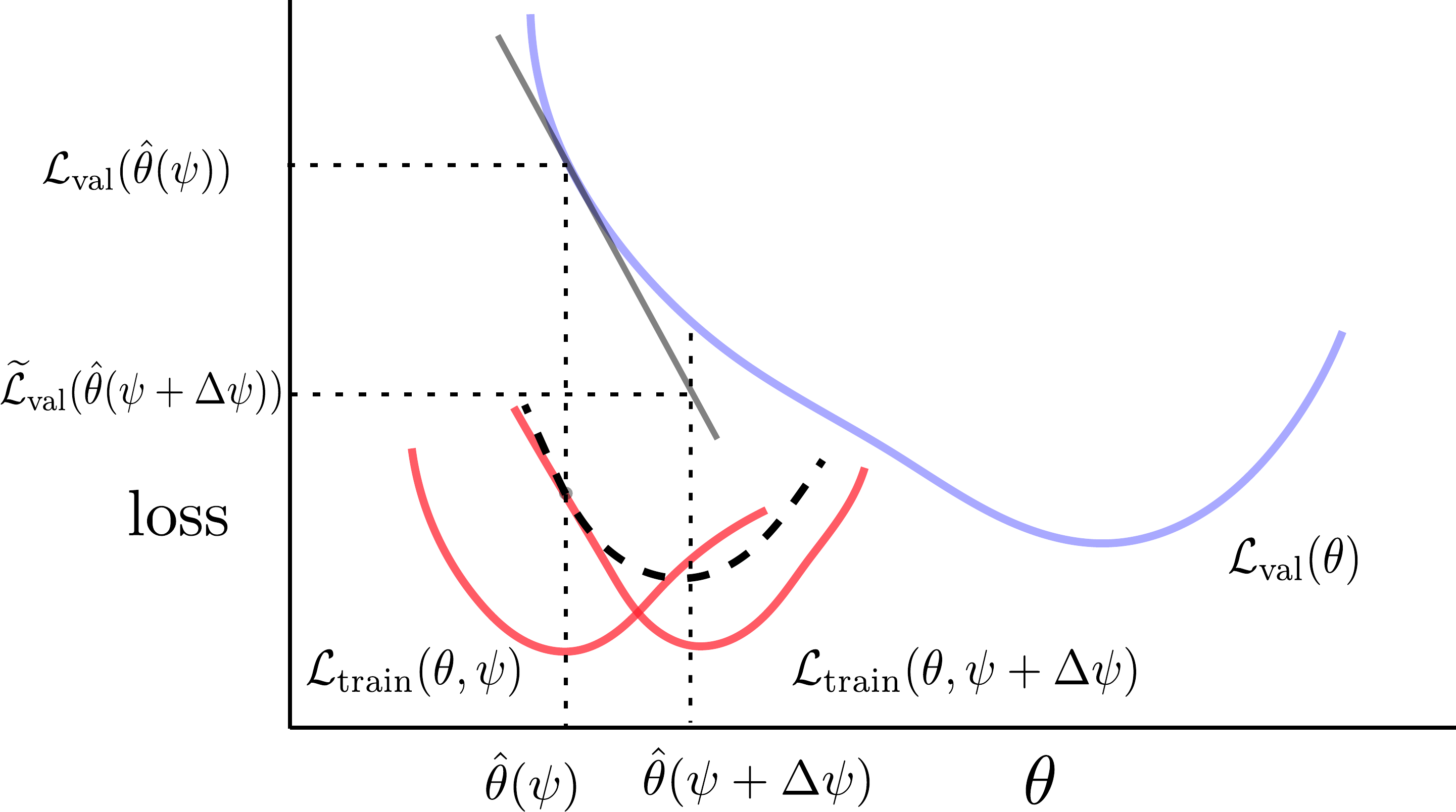}
    \caption{\textbf{Visualization of proposed differentiable approximation of objective $\sL_{\text{val}}(\hat{\bm{\theta}}({\bm\psi}))$} (blue). Red curves show the loss surface $\sL_{\text{train}}(\bm{\theta}, \bm{\psi})$ for the current training data and the loss surface $\sL_{\text{train}}(\bm{\theta}, \bm{\psi} + \Delta\bm{\psi})$ for data after a small update $\Delta \bm{\psi}$ in the simulator parameters. We assume $\hat{\bm{\theta}}(\bm{\psi} + \Delta\bm{\psi})$ to be close to $\hat{\bm{\theta}}(\bm{\psi})$, and use a single step of Newton's method to update $\bm{\theta}$. This gives us an approximation for updates in optimal $\hat{\bm{\theta}}$. We then use these to construct an approximation (black) for updates in optimal $\bm{\psi}$. \HB{Improve figure by making the font sizes larger and lines thicker}}
    \label{fig:visualize_method}
\end{figure}

\textbf{Differentiable Approximation}
Putting this back into Equation~\ref{eq:taylor_val}, and ignoring higher-order terms in $\Delta\bm{\theta}$, we get the approximation for our objective as
\begin{align} \label{eq:taylor_val_approx}
\widetilde{\sL}_{\text{val}}\big(\hat{\bm{\theta}}(\bm{\psi}_t + \Delta\bm{\psi})\big) = \sL_{\text{val}}\big(\hat{\bm{\theta}}(\bm{\psi}_t)\big) & \;- \nonumber\\
\frac{\partial \sL_{\text{train}}\big(\hat{\bm{\theta}}(\bm{\psi}_t), \bm{\psi}_t+ \Delta\bm{\psi}\big)}{\partial\bm{\theta}}^\top &\bm{H}\big(\hat{\bm{\theta}}(\bm{\psi}_t), \bm{\psi}_t+ \Delta\bm{\psi}\big)^{-1} \frac{d \sL_{\text{val}}\big(\hat{\bm{\theta}}(\bm{\psi}_t)\big)}{d\bm{\theta}}\;,
\end{align}
which is equivalent to writing
\begin{align} \label{eq:taylor_val_approx_eq}
\widetilde{\sL}_{\text{val}}\big(\hat{\bm{\theta}}(\bm{\psi})\big) &= \sL_{\text{val}}\big(\hat{\bm{\theta}}(\bm{\psi}_t)\big) - \frac{\partial \sL_{\text{train}}\big(\hat{\bm{\theta}}(\bm{\psi}_t), \bm{\psi}\big)}{\partial\bm{\theta}}^\top \bm{H}\big(\hat{\bm{\theta}}(\bm{\psi}_t), \bm{\psi}\big)^{-1}\;\frac{d \sL_{\text{val}}\big(\hat{\bm{\theta}}(\bm{\psi}_t)\big)}{d\bm{\theta}},
\end{align}
and is our local approximation of the objective. A visualization of this approximation is provided in Figure~\ref{fig:visualize_method}.

We propose to optimize our local model using gradient descent, and its derivative at point $\bm{\psi}_t$ (simulator parameter at iteration $t$) can be written as
\begin{align}
\frac{\partial \widetilde{\sL}_{\text{val}}\big(\hat{\bm{\theta}}(\bm{\psi})\big)}{\partial\bm{\psi}} \bigg|_{\bm{\psi}=\bm{\psi}_t}
\!\!\!\!\!\!\!\!\!\!\!=- \frac{\partial}{\partial\bm{\psi}} \bigg[\frac{\partial \sL_{\text{train}}\big(\hat{\bm{\theta}}(\bm{\psi}_t), \bm{\psi}\big)}{\partial\bm{\theta}} ^\top \!\!\!\bm{H}\big(\hat{\bm{\theta}}(\bm{\psi}_t), \bm{\psi}\big)^{-1} \frac{d \sL_{\text{val}}\big(\hat{\bm{\theta}}(\bm{\psi}_t)\big)}{d\bm{\theta}} \bigg]\!\bigg|_{\bm{\psi}=\bm{\psi}_t}\!\!\!\!.
\end{align}

Using the definition of $\sL_{\text{train}}\big(\hat{\bm{\theta}}(\bm{\psi}_t), \bm{\psi}\big)$ and ignoring the higher order derivative $\frac{\partial}{\partial\bm{\psi}}H\big(\hat{\bm{\theta}}(\bm{\psi}_t), \bm{\psi}\big)$, we get
\begin{align}
\frac{\partial \widetilde{\sL}_{\text{val}}\big(\hat{\bm{\theta}}(\bm{\psi})\big)}{\partial\bm{\psi}} \bigg|_{\bm{\psi}=\bm{\psi}_t}
\!\!\!\!\!\!\!\!\!\!=- \frac{\partial}{\partial\bm{\psi}} \!\!\mathop{\mathbb{E}}_{\zeta \sim p_{\psi}} \!\!\Big[\frac{\partial}{\partial\bm{\theta}}l\big(\zeta,\hat{\bm{\theta}}(\bm{\psi}_t)\big)\Big] ^\top\bigg|_{\bm{\psi}=\bm{\psi}_t} \!\!\!\!\!\!\!\!\bm{H}\big(\hat{\bm{\theta}}(\bm{\psi}_t), \bm{\psi}_t\big)^{-1} \frac{d}{d\bm{\theta}} \sL_{\text{val}}\big(\hat{\bm{\theta}}(\bm{\psi}_t)\big).
\end{align}

Next we show how to approximate the term $\frac{\partial}{\partial\bm{\psi}} \mathop{\mathbb{E}}_{\zeta \sim p_{\psi}} \Big[\frac{\partial}{\partial\bm{\theta}}l\big(\zeta,\hat{\bm{\theta}}(\bm{\psi}_t)\big)\Big] \in \mathbb{R}^{m\times n}$ which requires backpropagation through the dataset generation.

\subsection{Stochastic Simulator (Data Generating Distribution)}

\begin{wrapfigure}{r}{0.25\linewidth}
\resizebox{0.95\linewidth}{!}{
  \centering
  \tikz{ %
    \node[latent] (psi) {$\psi$} ; %
    \node[latent, right=of psi] (s) {s} ; %
    \node[det, right=of s] (zeta) {$\zeta$} ; %
    \plate[inner sep=0.25cm, xshift=-0.12cm, yshift=0.12cm] {plate1} {(s) (zeta)} {N}; %
    \edge {psi} {s} ; %
    \edge {s} {zeta} ; %
  }}
\end{wrapfigure}
We assume a stochastic simulator that involves a deterministic renderer, which may be non-differentiable, and we make the stochasticity in the process explicit by separating the stochastic part of the simulator from the deterministic rendering. Given the deterministic renderer component $\zeta = \mathrm{r}(s)$, we would like to find the optimal values of simulator parameters $\bm{\psi}$ that parameterize $s \sim q_{\bm{\psi}}(s)$ representing the stochastic component, expressing the overall simulator as $\zeta \sim p_{\bm{\psi}}(\zeta)$. Thus we can write
\begin{align}
p_{\bm{\psi}}(\zeta) &=\int_{s \in \{s|r(s)=\zeta\}} q_{\bm{\psi}}(s) ds\;.
\end{align}
For example, lets say we want to optimize the location of an object in a scene. Then $\psi$ could be the parameters of a Gaussian distribution $q_{\psi}(.)$ that is used to sample the location of the object in world coordinates, and $s$ denotes the location of the object sampled as $s \sim q_{\psi}(s)$. Now this sampled location $s$ is given as input to the renderer to generate an image $\zeta$ as $\zeta = \mathrm{r}(s)$. The overall simulator, including the stochastic sampling and the deterministic renderer, thus samples the images $\zeta$ as $\zeta \sim p_{\psi}(\zeta)$, where $p_{\psi}(\zeta)$ denotes the distribution over the images, parameterized by $\psi$.
Therefore we get
\begin{align}
\frac{\partial}{\partial\bm{\psi}} \mathop{\mathbb{E}}_{\zeta \sim p_{\bm{\psi}}} \Big[\frac{\partial}{\partial\bm{\theta}}l\big(\zeta,\hat{\bm{\theta}}(\bm{\psi}_t)\big)\Big]
&= \frac{\partial}{\partial\bm{\psi}} \mathop{\mathbb{E}}_{s \sim q_{\bm{\psi}}} \Big[\frac{\partial}{\partial\bm{\theta}}l\big(r(s),\hat{\bm{\theta}}(\bm{\psi}_t)\big)\Big].
\end{align}

The gradient of expectation term for continuous distributions can be computed using REINFORCE \cite{Williams_1992} as
\begin{align}
\frac{\partial}{\partial\bm{\psi}} \mathop{\mathbb{E}}_{s \sim q_{\bm{\psi}}} \Big[\frac{\partial}{\partial\bm{\theta}}l\big(r(s),\hat{\bm{\theta}}(\bm{\psi}_t)\big)\Big] 
&= \mathop{\mathbb{E}}_{s \sim q_{\bm{\psi}}} \left[\frac{d}{d\bm{\psi}} \log q_{\bm{\psi}}(s). \Big[\frac{\partial}{\partial\bm{\theta}}l\big(r(s),\hat{\bm{\theta}}(\bm{\psi}_t)\big)\Big]^\top \right] \nonumber \\
&\approx \sum_{k=1}^{K} \frac{d}{d\bm{\psi}} \log q_{\bm{\psi}}(s_k). \Big[\frac{\partial}{\partial\bm{\theta}}l\big(r(s_k),\hat{\bm{\theta}}(\bm{\psi}_t)\big)\Big]^\top,
\end{align}

where $s_k$ denotes samples drawn from the distribution $q_{\bm{\psi}}$.
This can be derived similarly for discrete distributions \cite{schulman2015gradient,rezende2014stochastic}, and we provide a derivation in the supplementary material.
Therefore we can write the update rule as
\begin{align}
\bm{\psi}_{t+1} &\leftarrow 
\bm{\psi}_t + \alpha\frac{\partial}{\partial\bm{\psi}} \mathop{\mathbb{E}}_{\zeta \sim p_{\psi}} \Big[\frac{\partial}{\partial\bm{\theta}}l\big(\zeta,\hat{\bm{\theta}}(\bm{\psi}_t)\big)\Big] ^\top\bigg|_{\bm{\psi}=\bm{\psi}_t} \!\!\bm{H}\big(\hat{\bm{\theta}}(\bm{\psi}_t), \bm{\psi}_t\big)^{-1} \frac{d}{d\bm{\theta}} \sL_{\text{val}}\big(\hat{\bm{\theta}}(\bm{\psi}_t)\big).
\label{eq:single_step_psi_update}
\end{align}

\begin{algorithm}[t]
	\caption{{AutoSimulate}}
	\label{algo:autosimulate}
	{
	\begin{algorithmic}
		\FOR {number of iterations} \let\labelitemi\labelitemii
		\STATE Sample dataset of size K: $D_\mathrm{train} \sim p_{\bm{\psi_t}}(\zeta)$
		\STATE Fine-tune model for $\epsilon$ epochs on $D_\mathrm{train}$
		\STATE Compute $\bm{H}\big(\hat{\bm{\theta}}(\bm{\psi}_t), \bm{\psi}_t\big)^{-1}\; \frac{d}{d\bm{\theta}} \sL_{\text{val}}\big(\hat{\bm{\theta}}(\bm{\psi}_t)\big)$ using CG
		\STATE Compute gradient of expectation as
		$\sum_{k=1}^{K}\!\! \frac{d}{d\bm{\psi}} \log q_{\bm{\psi}}(s_k). \Big[\frac{\partial}{\partial\bm{\theta}}l\big(r(s_k),\hat{\bm{\theta}}(\bm{\psi}_t)\big)\Big]^\top$
		\STATE Update simulator by descending the gradient $-\frac{\partial}{\partial\bm{\psi}} \mathop{\mathbb{E}}_{\zeta \sim p_{\psi}} \Big[\frac{\partial}{\partial\bm{\theta}}l\big(\zeta,\hat{\bm{\theta}}(\bm{\psi}_t)\big)\Big] ^\top\bigg|_{\bm{\psi}=\bm{\psi}_t} \bm{H}\big(\hat{\bm{\theta}}(\bm{\psi}_t), \bm{\psi}_t\big)^{-1}\; \frac{d}{d\bm{\theta}} \sL_{\text{val}}\big(\hat{\bm{\theta}}(\bm{\psi}_t)\big)$
		\ENDFOR
	\end{algorithmic}
	}
\end{algorithm}

It can be seen that we have transformed our original bi-level objective in Equation~\ref{eq:main_obj_psi}, into iteratively creating and minimizing a local model $\widetilde{\sL}_{\text{val}}(\hat{\bm{\theta}}\big(\bm{\psi}_t + \Delta\bm{\psi})\big)$. %
An overview of the method can be found in algorithm \ref{algo:autosimulate}.

\subsection{Efficient Numerical Computation}

The benefit of the proposed approximation is that it enables us to use techniques from unconstrained optimization. The update rule in Equation~\ref{eq:single_step_psi_update} requires an inverse Hessian computation at each iteration, which is common in second-order optimization. We now discuss an efficient strategy for optimizing our model.

\textbf{Regularization for Hessian}
The first challenge is that the Hessian might have negative eigenvalues. Thus the inverse of the Hessian may not exist. We regularize the Hessian using the Levenberg method \cite{more1978levenberg} and use $\bm{H} + \lambda \bm{I}$, where $\lambda$ is the regularization constant and $\bm{I}$ denotes the identity matrix. This is common in second-order optimization of Neural Networks \cite{pmlr-v70-botev17a,henriques2018small,martens2015optimizing}.

\textbf{Inverse Hessian--vector product computation}
Secondly, to compute the update term in Equation~\ref{eq:single_step_psi_update}, we split it as follows: we first compute $\bm{v}_{\bm{\psi}_t} \eqdef \bm{H}_{\hat{\bm{\theta}}(\bm{\psi}_t)}^{-1} \bm{g}_{\textrm{val}}$ and then compute $\frac{\partial}{\partial\bm{\psi}} \widetilde{\sL}_{\text{val}}\big(\hat{\bm{\theta}}(\bm{\psi})\big) \bigg|_{\bm{\psi}=\bm{\psi}_t} = - \bm{v}_{\bm{\psi}_t}. \nabla_{\bm{\psi}}\, \bm{g}_\textrm{train}$, where $\bm{g}_\textrm{val} \eqdef \frac{d}{d\bm{\theta}} \sL_{\textrm{val}}\big(\hat{\bm{\theta}}(\bm{\psi}_t)\big)$ and $\bm{g}_\textrm{train} \eqdef \frac{d}{d\bm{\theta}} \sL_{\text{train}}\big(\hat{\bm{\theta}}(\bm{\psi}_t), \bm{\psi}\big)$.

The inverse Hessian vector product $\bm{H}^{-1}\bm{g}$ is computed by using the conjugate gradient method \cite{10.5555/865018} to solve $ \min_{\bm{v}} \{\inv{2} \bm{v}^\top \bm{H}\bm{v} - \bm{g}^\top \bm{v}\}$. This is common in second-order optimization. Thus the update term in Equation~\ref{eq:single_step_psi_update} can be obtained as
\begin{subequations}
\begin{align}
\bm{v}_{\bm{\psi}_t} &\equiv \arg\min_{\bm{v}} Q(\bm{v}) \eqdef \{\inv{2} \bm{v}^\top \bm{H}_{\hat{\bm{\theta}}(\bm{\psi}_t)} \bm{v} -  \bm{g}_{\textrm{val}}^\top \bm{v}\}\;, \label{eq:cg}\\
\bm{\psi}_{t+1} &\leftarrow \bm{\psi}_t + \alpha\,\bm{v}_{\bm{\psi}_t}. \nabla_{\bm{\psi}} \bm{g}_\textrm{train}\;.
\end{align}
\end{subequations}
The CG approach only requires the evaluation of $\bm{H}_{\hat{\bm{\theta}}}\bm{v}$. Using automatic differentiation, Hessian-vector product requires only one forward and backward pass, same as a gradient. In practise, a good approximation for the inverse hessian vector product can be obtained with few iterations.

\textbf{Approximations for $\Delta\hat\theta_{\psi}$} We proposed a novel approximation for the solution of the inner problem. To further reduce the compute overhead, we propose approximations for $\Delta\hat\theta_{\psi}$. Table \ref{tab:theta_approx} shows some other alternative approximations for the inner problem, which can be obtained by using a linear approximation for the inner problem or using an approximate quadratic approximation. 
Using automatic differentiation, Hessian-vector product requires only one forward and backward pass, same as a gradient.
Another baseline we try is a constant approximation for the inner problem where the method does not use the real validation set at all and just finds data which gives minimum model loss.
\begin{table}
    {\tiny
    \caption{\textbf{Proposed approximations for $\Delta\hat\theta_{\psi}$.} \HB{fix here}}
    \label{tab:theta_approx}
	\setlength{\tabcolsep}{-5pt}
	\renewcommand{\arraystretch}{1.2}
    \begin{tabularx}{\textwidth}{@{}p{1.5cm}p{4.3cm}X@{}}
    \toprule
                            & Approximation ($\Delta\hat\theta_{\psi}$) & Derivative Term ($\frac{\partial}{\partial\bm{\psi}} \widetilde{\sL}_{\text{val}}(\hat{\bm{\theta}}(\bm{\psi}))$) \\
    \midrule
        Quadratic   					
        & $-H(\hat{\bm{\theta}}(\bm{\psi}_t), \bm{\psi})^{-1} \!\frac{\partial}{\partial\bm{\theta}} \sL_{\text{tr.}}(\hat{\bm{\theta}}(\bm{\psi}_t), \bm{\psi})$ 
        &  $- \frac{\partial}{\partial\bm{\psi}} \mathop{\mathbb{E}}_{\zeta \sim p_{\psi}} \![\frac{\partial}{\partial\bm{\theta}}l(\zeta,\hat{\bm{\theta}}(\bm{\psi}_t))] ^\top\bigg|_{\bm{\psi}=\bm{\psi}_t}\!\!\!\!\!\!\!\!\! \bm{H}(\hat{\bm{\theta}}(\bm{\psi}_t), \bm{\psi}_t)^{-1} \frac{d}{d\bm{\theta}} \sL_{\text{val}}(\hat{\bm{\theta}}(\bm{\psi}_t))$\\
        Approx. Quadratic   		
        & $H(\hat{\bm{\theta}}(\bm{\psi}_t), \bm{\psi})\frac{\partial}{\partial\bm{\theta}} \sL_{\text{tr.}}(\hat{\bm{\theta}}(\bm{\psi}_t), \bm{\psi})$ 
        & $\frac{\partial}{\partial\bm{\psi}} \mathop{\mathbb{E}}_{\zeta \sim p_{\psi}} [\frac{\partial}{\partial\bm{\theta}}l(\zeta,\hat{\bm{\theta}}(\bm{\psi}_t))] ^\top\bigg|_{\bm{\psi}=\bm{\psi}_t} \!\!\!\!\!\! \bm{H}(\hat{\bm{\theta}}(\bm{\psi}_t), \bm{\psi}_t) \frac{d}{d\bm{\theta}} \sL_{\text{val}}(\hat{\bm{\theta}}(\bm{\psi}_t))$\\
        Linear       					& $-\frac{\partial}{\partial\bm{\theta}}\sL_{\text{tr.}}(\hat{\bm{\theta}}(\bm{\psi}_t), \bm{\psi})$ &  $- \frac{\partial}{\partial\bm{\psi}} \mathop{\mathbb{E}}_{\zeta \sim p_{\psi}} [\frac{\partial}{\partial\bm{\theta}}l(\zeta,\hat{\bm{\theta}}(\bm{\psi}_t))] ^\top\bigg|_{\bm{\psi}=\bm{\psi}_t} \frac{d}{d\bm{\theta}} \sL_{\text{val}}(\hat{\bm{\theta}}(\bm{\psi}_t))$\\
        No Val         			& \textbf{1} 						& $- \frac{\partial}{\partial\bm{\psi}} \mathop{\mathbb{E}}_{\zeta \sim p_{\psi}} [\frac{\partial}{\partial\bm{\theta}}l(\zeta,\hat{\bm{\theta}}(\bm{\psi}_t))] ^\top\bigg|_{\bm{\psi}=\bm{\psi}_t}$ \\
    \bottomrule
    \end{tabularx}}
\end{table}
\vspace{-0.1in}

\section{Experiments}
In this section, we demonstrate the effectiveness of the proposed method in learning simulator parameters in two different scenarios. First we evaluate our method on a simulator with the goal of performing a per-pixel semantic segmentation task. Second, we also conduct experiments with physically based rendering for solving object detection task on real world data.
In supplementary material we provide more details about the data generation process from a simulator. In our experiments, we have used two physically based simulators: Blender-based CLEVR and the Arnold renderer.

\textbf{Baselines}
In all our experiments, we compare our proposed method for learning simulator parameters against three state-of-the-art baseline algorithms.
The main baseline is ``learning to simulate'' (LTS) \cite{ruiz2018learning} which uses the REINFORCE gradient estimator. As the code for this is not public, we implemented it. Please note that Meta-sim \cite{kar2019metasim} also uses REINFORCE.
In addition, we also compare against the two most established hyper-parameter optimization algorithms in machine learning; for Bayesian optimization we use the opensource Python package \texttt{bayesian-optimization} \cite{python_bo} and for random search we used the Scikit-learn Python library \cite{scikit-learn}.
Please note that prior work \cite{ruiz2018learning,kar2019metasim,yang2019learning} did not compare against these two approaches and in our results we found that out-of-the-box BO and Random search outperform REINFORCE \cite{ruiz2018learning}.

\subsection{CLEVR Blender}
In this experiment, we use the CLEVR simulator \cite{clevr_cite} which generates physically based images. The main task is semantic segmentation of the three classes present in the CLEVR benchmark, namely, Sphere, Cube and Cylinder. The images are generated using the CLEVR dataset generator. We optimize multiple CLEVR rendering parameters including the intensity of ambient light, back light, number of samples, number of bounces of light, image size, location of the objects in the scene, and materials of objects. %
The validation set is composed of synthetic images generated with a particular simulator configuration shown in Figure \ref{fig:clevr_val}. 

\textbf{Task Network} For the task network, we use a UNet \cite{ronneberger2015u} with eight convolutional layers. UNet is very common for segmentation and we have used an openly available Pytorch implementation\footnote{\url{https://github.com/jvanvugt/pytorch-unet}} of UNet. The performance is measured in terms of mean IoU (intersection over union).%

\textbf{Results} Quantitative results are shown in Table \ref{tab:clevr_segmentation_quantitative}.
BO and LTS methods to learn simulator parameters achieve similar test accuracy to ours. However, both these methods generate significantly more images to reach a similar accuracy. Essentially, to reach to the same level of test accuracy, the proposed approach requires $2.5\times$ and $5\times$ less data than BO and LTS methods respectively. This translates into saving time and resources required for the data generation and CNN training steps. Figure~\ref{fig:clevr_val} shows some qualitative examples for this task.
\begin{table}
    \centering
    \caption{\textbf{Segmentation on Clevr.} Comparison of time, number of images, and test accuracy achieved by different methods.}
    \label{tab:clevr_segmentation_quantitative}
    \scriptsize
    \begin{tabularx}{0.7\textwidth}{Xrrr}
    \toprule
        Method              & Time 	& Images 	& Test mIoU\\
    \midrule
        REINFORCE (LTS)     &  3h2m 	&  3,750  	& 61.0 \\
        Random search       &  2h38m	&  2,090  	& 60.8 \\
        BO                	&  \textbf{43m} 		&  960  	& 62.9 \\
        AutoSimulate        &  53m 		&  \textbf{390}  	& \textbf{62.9} \\
    \bottomrule
    \end{tabularx}
\end{table}

\subsection{Photorealistic Renderer Arnold}
We next evaluate the performance of our proposed method on real-world data. For this, we use LineMod-Occluded (LM-O) dataset \cite{hodan2018bop} for object detection task that consists of 3D models of objects. The dataset consists of eight object classes that includes metallic, non-Lambertian objects, e.g., metallic cans. The data has recently been used for benchmarking object detection problem \cite{hodan2019photorealistic}. We use the same test split for evaluating the performance of our method. Further, we use the same simulator as Hodan et al. \cite{hodan2019photorealistic}, based on Arnold \cite{georgiev2018arnold}, to generate photo-realistic synthetic data for training an object detector model.
Note that Hodan et al. \cite{hodan2019photorealistic} heavily relied on human expert knowledge to correctly decide the distributions for different simulator parameters.
In comparison, we show how our approach can be used to instead learn the optimal distribution over the simulator parameters without sacrificing accuracy.

In this experiment, we are given three scenes and nine locations within each scene. These locations signify the locations within the scene where objects can be placed. They can be arbitrarily chosen or can be selected by a human. Further, there are two rendering quality settings (high and low).
The task is to optimize the categorical distribution for finding the fraction of data to be generated from each of these locations from different scenes under the two quality settings. Thus, the problem requires optimising 54 simulator parameters. Some of the images generated during simulator training have been shown in Figure~\ref{fig:obj_det_sample_images}. 

\def \imwidth {0.24}
\begin{figure*}
	\centering
	\resizebox{\textwidth}{!}{
	\begin{tabular}{cccc}
		\includegraphics[width=\imwidth\textwidth]{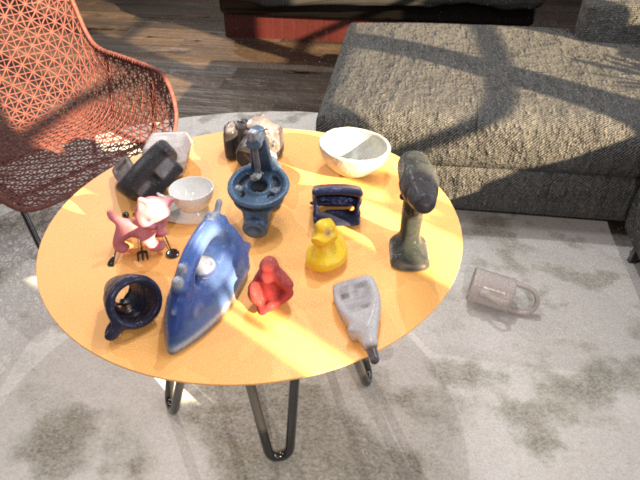}
		&
		\includegraphics[width=\imwidth\textwidth]{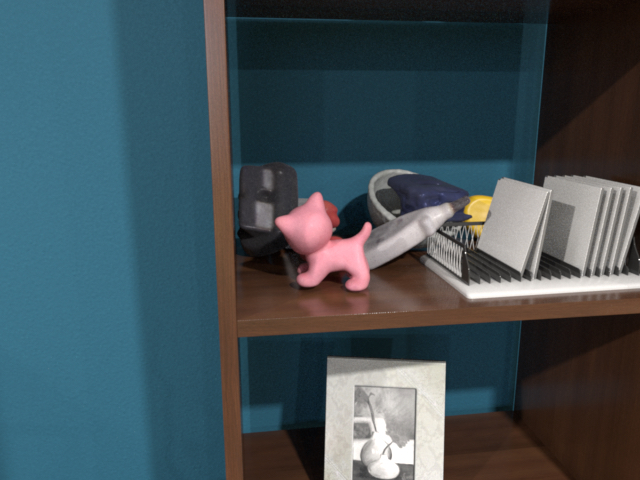}
		&
		\includegraphics[width=\imwidth\textwidth]{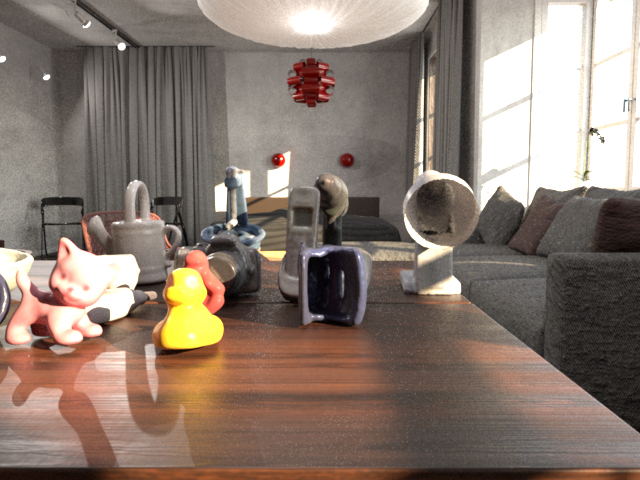}
		&
		\includegraphics[width=\imwidth\textwidth]{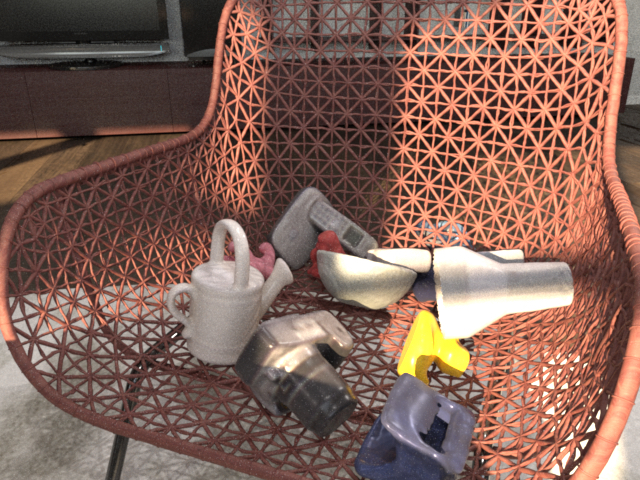}\\
		\includegraphics[width=\imwidth\textwidth]{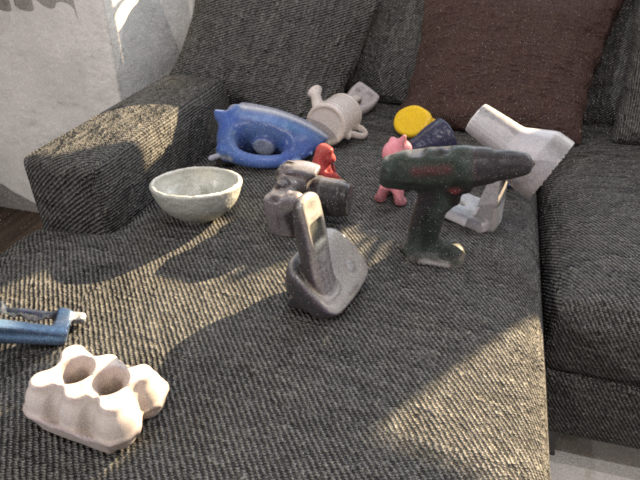}
		&
		\includegraphics[width=\imwidth\textwidth]{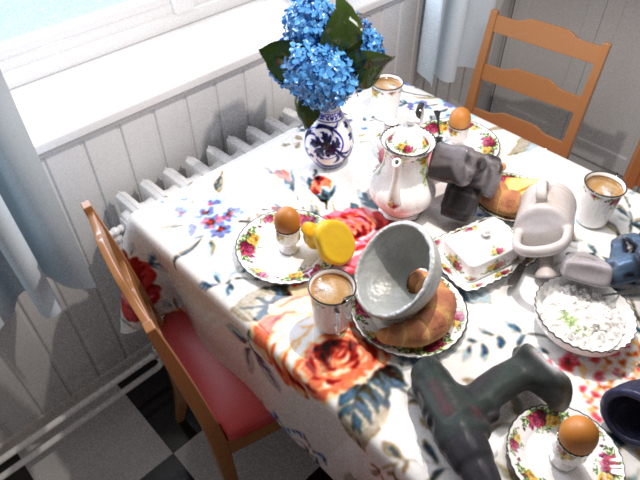}
		&
		\includegraphics[width=\imwidth\textwidth]{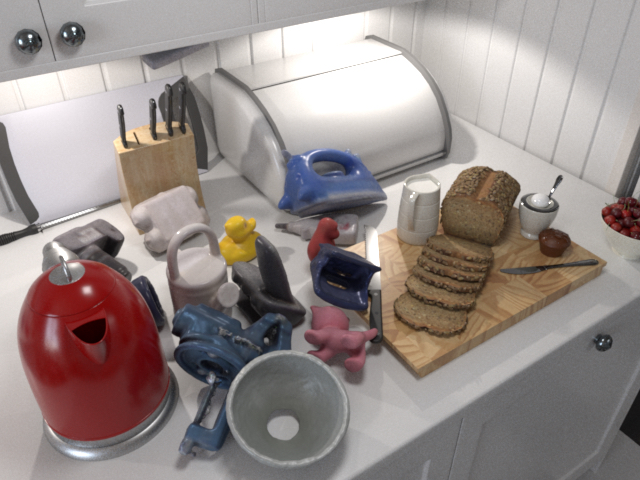}
		&
		\includegraphics[width=\imwidth\textwidth]{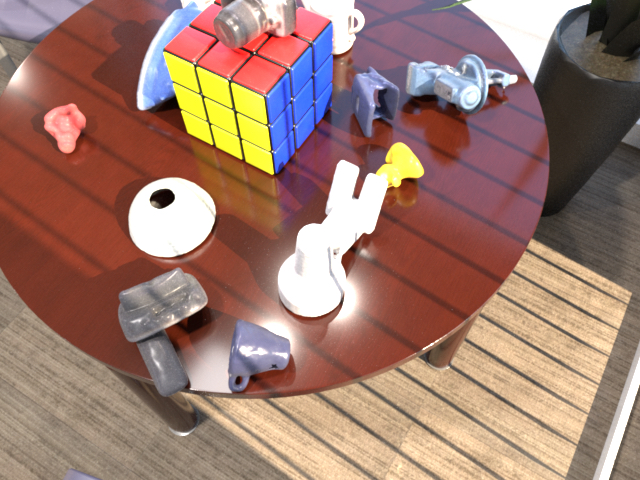}

		\\
	\end{tabular}}
	\caption{\textbf{Synthetic images generated with Arnold renderer} used for training.}
	\label{fig:obj_det_sample_images}
\end{figure*}

\textbf{Task Network} For this task we use Yolo \cite{redmon2017yolo9000}, which is an established method for object detection and the code is freely available and easy to use\footnote{\url{https://github.com/ultralytics/yolov3}}. We use Yolo-spp which has 112 layers with default parameter setting. The object detection performance is measured in terms of mean average precision (mAP@0.50).%

\textbf{Results}
\begin{table}
    \caption{\textbf{Object Detection on real-world dataset LM-O.} Comparison of methods. We run each method for a 1,000 epochs and report: \emph{Val. mAP}: maximum validation mAP, \emph{Images} and \emph{Time}: number of images generated and time spent to reach maximum validation mAP, \emph{Test mAP}: test mAP of the result.}
    \label{tab:yolo_detection_quantitative}
    \centering
    \scriptsize
    \begin{tabularx}{0.8\textwidth}{Xrrrr}
    	\toprule
        Method                                      & Val. mAP 		& Images 	& Time(s) & Test mAP\\
        \midrule
        REINFORCE (LTS)                              & 40.2 	 		&  86,150	&  114,360 	&  37.2 \\
        Bayesian Optimization                 		& 39.3 	 		&  9,200		&  83,225 &  37.5\\ 
        Random Search                               & 40.3  			&  34,300		&  134,318 	&  37.0\\
        AutoSimulate &37.1 & 8,950 & 23,193 & 36.1\\
        AutoSimulate(Approx Quad)                    & 40.1 	 			&  \textbf{2,950}			&  \textbf{2,321} &  37.4\\
        AutoSimulate (Linear)                         & \textbf{41.4} 	 		&  17,850	&  30,477 &  \textbf{45.9}\\
    \bottomrule
    \end{tabularx}
\end{table}
Quantitative results are provided in Table~\ref{tab:yolo_detection_quantitative} where we compare the presented method against the baselines. 
We evaluate these methods on three different criteria: mAP accuracy achieved on the object detection test set, total images generated during training of the simulator, and total time taken to complete simulator training.
AutoSimulate provides significant benefit over the baseline methods on all the criteria. Our method, AutoSimulate (Linear), achieves a remarkable improvement of almost $8$ percent in mAP on test set. Further, it requires much lesser data (Figure \ref{fig:data_gen_histogram}) generation in comparison to baseline methods, and takes almost 2.5--4$\times$ less time to train simulator parameters compared to all the baselines including LTS, BO and random search. Our AutoSimulate (Approx Quad) is almost 35--60$\times$ faster than the baselines while also achieving the same mAP accuracy. This shows the effectiviness of the proposed method for learning simulator parameters. In Figure~\ref{fig:obj_det_test_images}, we show qualitative examples of object detections in real-world images from the LM-O dataset, using a neural network trained on synthetic images generated by Arnold renderer optimized using AutoSimulate.
\def \imwidth {0.32}
\begin{figure*}
	\centering
	\resizebox{\textwidth}{!}{
	\begin{tabular}{cccc}
		\includegraphics[width=\imwidth\textwidth]{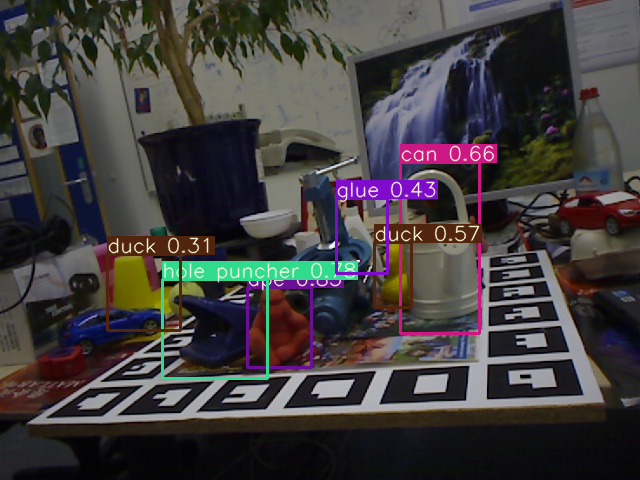}
		&
		\includegraphics[width=\imwidth\textwidth]{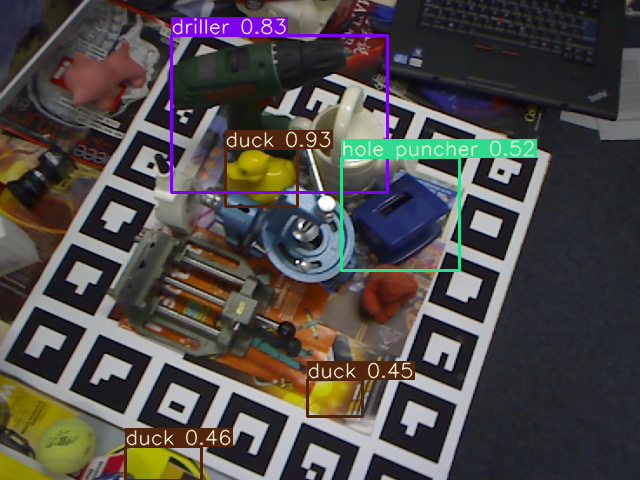}
		&
		\includegraphics[width=\imwidth\textwidth]{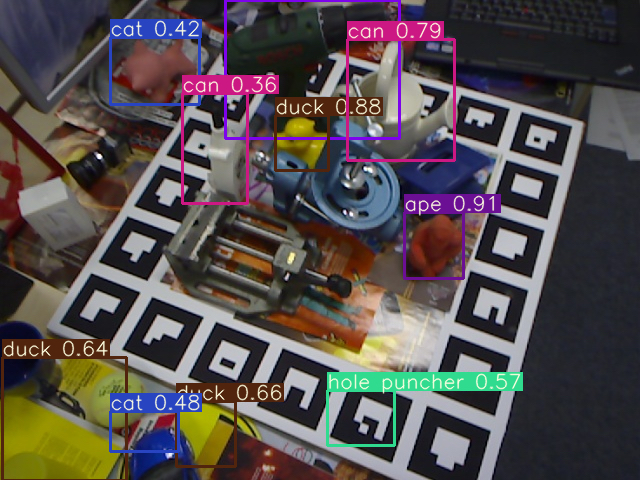}
		&
		\includegraphics[width=\imwidth\textwidth]{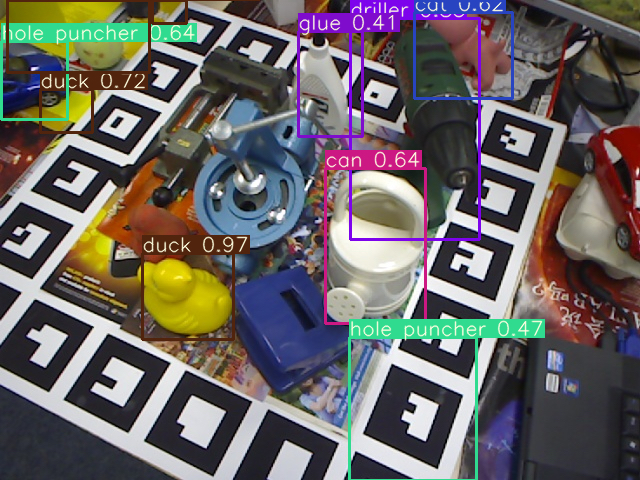}\\
		\includegraphics[width=\imwidth\textwidth]{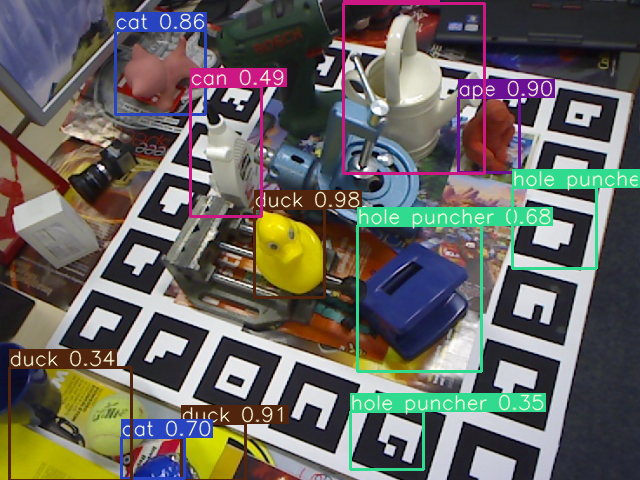}
		&
		\includegraphics[width=\imwidth\textwidth]{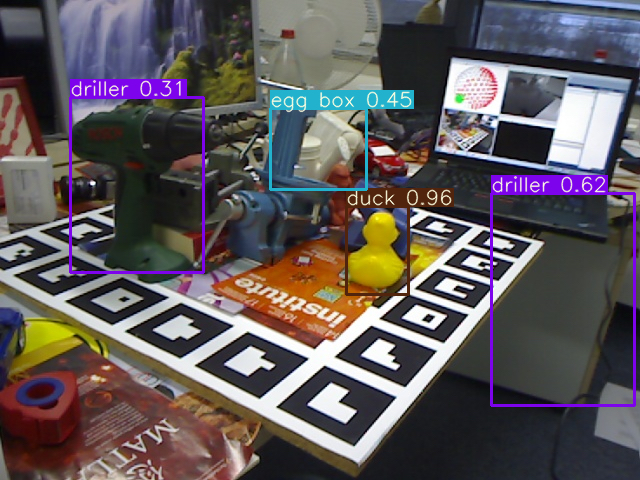}
		&
		\includegraphics[width=\imwidth\textwidth]{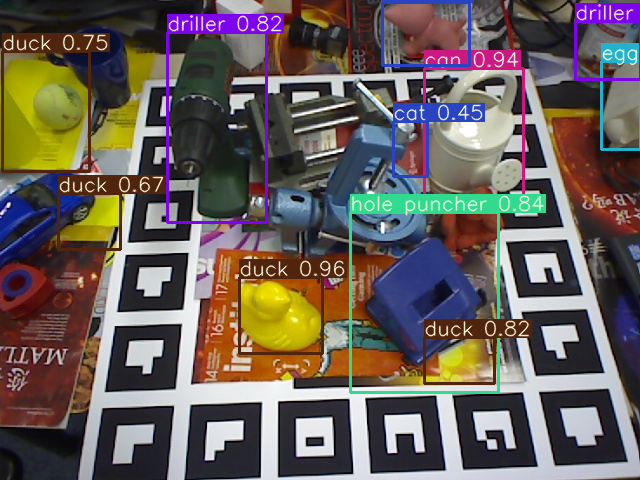}
		&
		\includegraphics[width=\imwidth\textwidth]{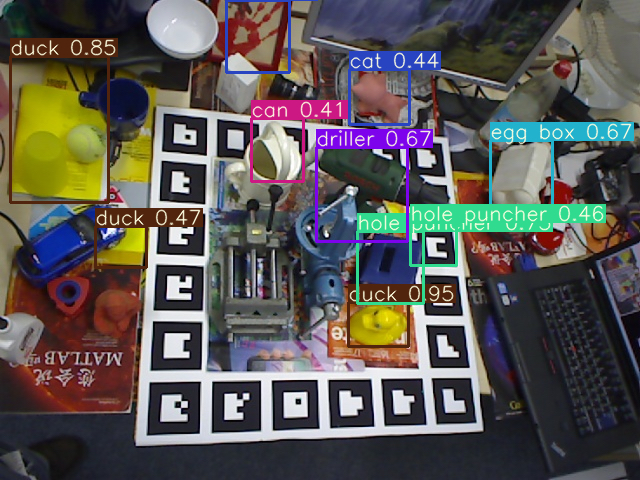}
		\\
	\end{tabular}}
	\caption{\textbf{Sample detections on real images from LM-O dataset}, using a network trained on synthetic images generated by Arnold renderer, which is optimized with AutoSimulate.}
	\label{fig:obj_det_test_images}
\end{figure*}

In supplementary material, we also show results on training simulator along with Faster-rcnn \cite{ren2017faster} another popular object detection model.

\def \imwidth {0.8}
\begin{figure}
  \checkoddpage
  \edef\side{\ifoddpage l\else r\fi}%
  \makebox[\textwidth][\side]{%
    \begin{minipage}[]{0.48\textwidth}
    \centering
	\begin{tabular}{cccccc}
	\includegraphics[trim=390 0 520 0,clip,width=\imwidth\textwidth]{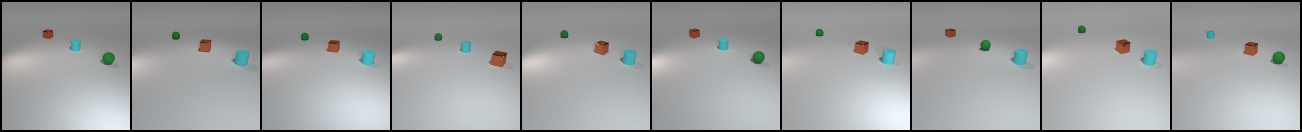}\\
	\includegraphics[trim=0 0 910 260,clip,width=\imwidth\textwidth]{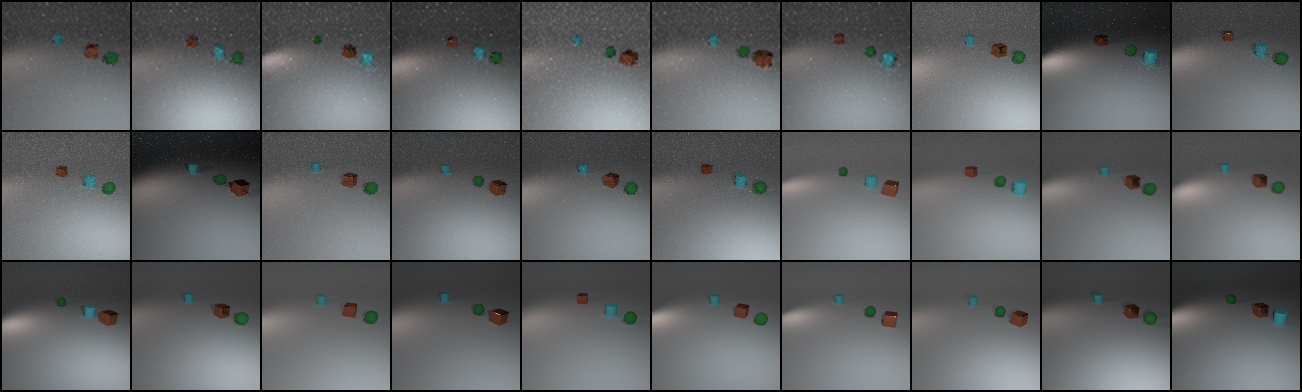}\\
	\end{tabular}
	\caption{\textbf{Images Rendered with the Clevr Simulator.} Top: samples from validation set. Bottom: images rendered during the simulator training, showing variation in the quality of images, lighting in the scene, and location of objects.}
	\label{fig:clevr_val}
    \end{minipage}%
    \hfill
    \begin{minipage}[]{0.49\textwidth}
      \centering
      \includegraphics[width=\textwidth]{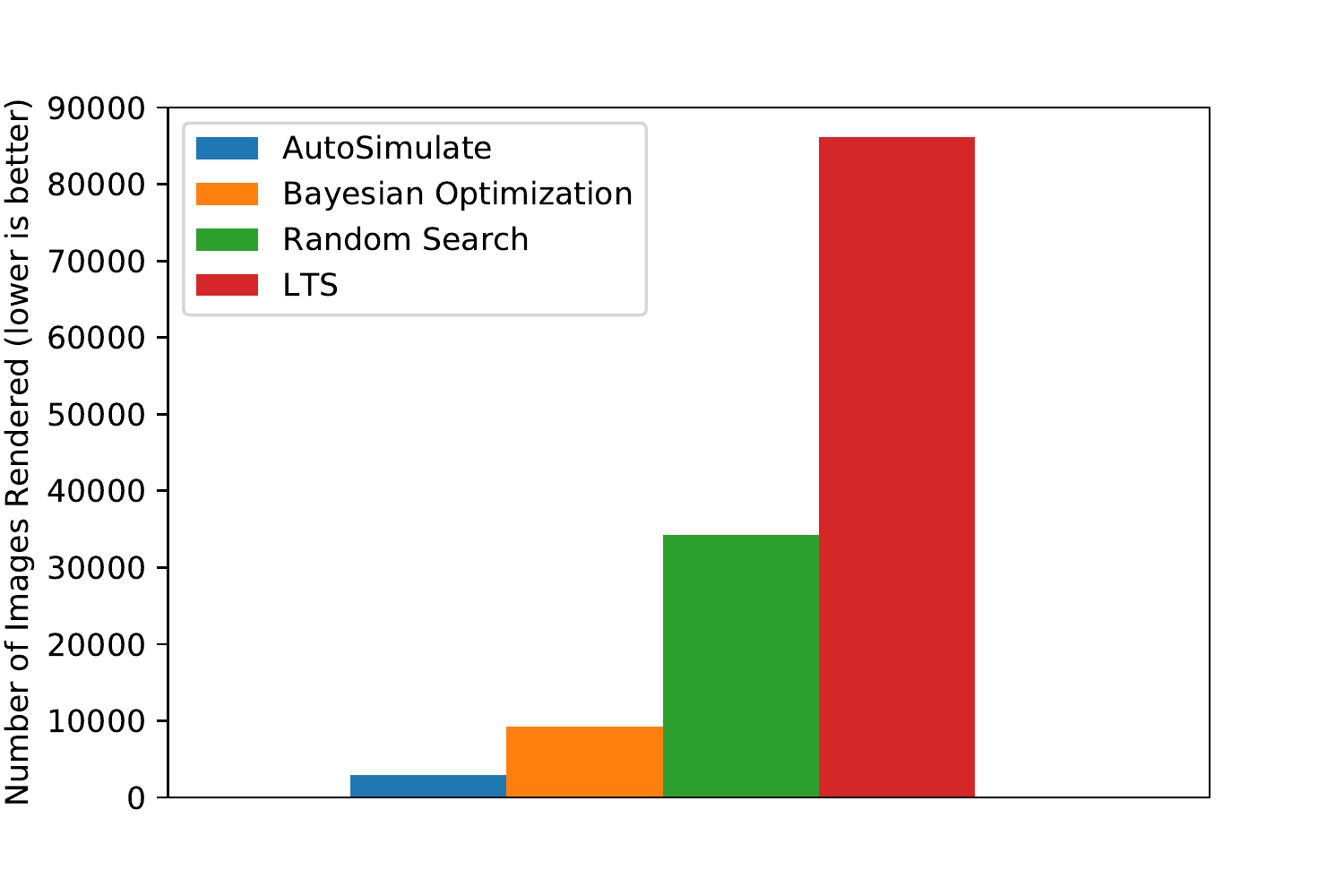}
      \caption{\textbf{Number of synthetic images required during training} using the photorealistic renderer Arnold with different methods.}
      \label{fig:data_gen_histogram}
    \end{minipage}%
  }%
\end{figure}

\subsection{Additional Studies}

\textbf{Approximations for $\theta$}
In this ablation, we analyse the different possible approximations for $\Delta\hat{\theta}$. The quantitative results are provided in Table \ref{tab:theta_approx_quantitative}. We observe that linear approximation of $\Delta\hat{\theta}$ achieves the best test accuracy (mAP). Further, our Approximate Quadratic takes the least time to converge and requires the least amount of data generation. Thereby giving the user freedom to select the approximation based on their speed--accuracy requirements.

\begin{table}
    \caption{\textbf{Effect of Approximations}}
    \label{tab:theta_approx_quantitative}
    \centering
    \scriptsize
    \begin{tabularx}{0.66\textwidth}{Xrrr}
    	\toprule
        Method                                      & Test mAP 			& Time(s) 			& Images\\
        \midrule
        Exact Quadratic (Ours)                      & 36.1 				&  2,3193 				&  8,950\\
        Approximate Quadratic (Ours)               & 37.4  			&  \textbf{2,321} 	&  \textbf{2,950}\\
        Linear  (Ours)                              & \textbf{45.9} 	&  30,477 			&  1,7850\\
        No Validation                    		& 29.3 				&  5,539 			&  6,400\\
    \bottomrule
    \end{tabularx}
\end{table}

\textbf{Effect of Freezing Layers}
It is a common practise to train on synthetic data with the initial layers of the network frozen and trained on real data. For this ablation, we use networks pretrained on COCO dataset. The effect of freezing different numbers of layers are shown in Table \ref{tab:obj_freezing_layers}. In particular, we show the effect of freezing 0, 98 and 104 layers out of the total 112 layers. We observe that freezing no layers achieves better accuracy than freezing layers of the CNN model. However, it leads to higher convergence time. The faster convergence of frozen layers can be attributed to fast Hessian approximation computation. %

\begin{table}
    \caption{\textbf{Effect of Freezing Layers}}
    \label{tab:obj_freezing_layers}
    \centering
    \scriptsize
    \begin{tabularx}{\textwidth}{X rrr@{\hskip 4mm}rrr@{\hskip 4mm}rrr}
    	\toprule
        & \multicolumn{3}{l}{0 frozen layers} & \multicolumn{3}{l}{98 frozen layers} & \multicolumn{3}{l}{104 frozen layers}\\
        \midrule
         Method              						& mAP 	& Time(s) 	& Images 		& mAP 	& Time(s) 	& Images 			& mAP 	& Time(s) & Images \\
        \midrule
        REINFORCE (LTS)                             	& 37.2 		&  114,360 	&  86,150 		& 33.0 	&  114,360 		&  86,150		& 31.9 	&  145,193 		&  104,600\\
        Bayesian Optimization                 		& 37.5 		&  83,225 	&  \textbf{9,225} 		& 31.7 	&  13,940 	&  3,550				& 31.7 	&  30,538 		&  6,050\\
        Random Search                               & 36.8 		&  134,137 	&  34,300		& 30.2 	&  8,913 	&  3,500 			& 28.9 	&  73,411 		&  21,650\\
        Ours                                        & \textbf{45.9} 		&  \textbf{30,477} 	&  17,850 		& \textbf{37.1} 	&  \textbf{2,321} 		&  \textbf{2,950}				& \textbf{35.8} 	&  \textbf{958} 		&  \textbf{1,000}\\
    \bottomrule
    \end{tabularx}
\end{table}

\textbf{Generelization and Effect of Network Size} 
In this ablation, we study whether a simulator trained on a shallow network generalizes to a deeper network.
We first examine the effect of network depth on simulator training. In particular, we show results of using two networks: YOLO-spp with 112 layers and YOLO-tiny with 22 layers in Table \ref{tab:ablation_network_depth}. %
Our approach on shallow network takes almost 7$\times$, 15$\times$, 135$\times$ less time to converge than LTS, random search and BO methods respectively. On the other hand, our method on deep network takes 4$\times$, 2.5$\times$ and 4$\times$ less time than the three baseline methods. This highlights that the relative improvement of our method with the shallow network is much better than the deeper network. Further, in the supplementary material we also show the generalization of simulator parameters trained using shallow network on generating data for training deeper network. It gives users freedom to select size of network according to resources available for training the simulator.
\begin{table}
    \caption{\textbf{Effect of Network Size}}
    \label{tab:ablation_network_depth}
    \centering
    \scriptsize
    \begin{tabularx}{0.75\textwidth}{Xrrr@{\hskip 4mm}rrr}
    	\toprule
        								& \multicolumn{3}{l}{Yolo-spp} 				& \multicolumn{3}{l}{Yolo-Tiny}\\
        \midrule
        Method              			& mAP 	& Time(s) 	& Images 		& mAP 		& Time(s) 	& Images\\
        \midrule
        REINFORCE (LTS)                  & 37.2 		&  114,360 	&  86,150		& \textbf{24.7} 		&  3,475 	&  11,550\\
        Bayesian Optimization           & 37.5 		&  83,225 	&  \textbf{9,225}			& 19.5 		&  65,760 	&  35,700\\
        Random Search                   & 36.8 		&  134,137 	&  34,300		& 20.6 		&  7,319 	&  11,620\\
        Ours                            & \textbf{45.9} 		&  \textbf{30,477} 	&  17,850		& 21.2 		&  \textbf{484} 		&  \textbf{280}\\
        \bottomrule
    \end{tabularx}
\end{table}

\section{Conclusion}
Recent methods optimize simulator parameters with the objective of maximising accuracy on a downstream task. However these methods are computationally very expensive which has hindered the widespread use of simulator optimization for generating optimal training data. In this work, we propose an efficient algorithm for optimally generating synthetic data, based on a novel differentiable approximation of the objective. We demonstrate the effectiveness of our approach by optimising state-of-the-art photorealistic renderers using a real-world validation dataset, where our method significantly outperforms previous methods.

\textbf{Acknowledgements}
Harkirat is wholly funded by a Tencent grant. This work was supported by ERC grant ERC-2012-AdG 321162-HELIOS, EPSRC grant Seebibyte EP/M013774/1 and EPSRC/MURI grant EP/N019474/1. We would like to acknowledge the Royal Academy of Engineering, and also thank Ondrej Miksik, Tomas Hodan and Pawan Mudigonda for helpful discussions. 

\bibliographystyle{splncs04}
\bibliography{main}
\end{document}